\documentclass[letterpaper]{article} 
\usepackage{aaai25}  
\usepackage{times}  
\usepackage{helvet}  
\usepackage{courier}  
\usepackage[hyphens]{url}  
\usepackage{graphicx} 
\usepackage{natbib}  
\usepackage{caption} 
\usepackage{algorithm}
\usepackage{algpseudocode}
\usepackage{amsmath}
\usepackage{multirow}
\usepackage{makecell}
\usepackage{placeins}
\usepackage{bbding}
\usepackage{newfloat}
\usepackage{listings}
\usepackage{amsfonts}

\DeclareCaptionStyle{ruled}{labelfont=normalfont,labelsep=colon,strut=off} 
\floatstyle{ruled}
\newfloat{listing}{tb}{lst}{}
\floatname{listing}{Listing}
\title{DocKylin: A Large Multimodal Model for Visual Document Understanding with Efficient Visual Slimming}
\author{
    Jiaxin Zhang\textsuperscript{\rm 1}\equalcontrib,
    Wentao Yang\textsuperscript{\rm 1}\equalcontrib,
    Songxuan Lai\textsuperscript{\rm 2},
    Zecheng Xie\textsuperscript{\rm 2, \Envelope},
    Lianwen Jin\textsuperscript{\rm 1, \Envelope}
}
\affiliations{
    \textsuperscript{\rm 1}South China University of Technology\\
    \textsuperscript{\rm 2}Huawei Cloud\\
    xiezecheng1@huawei.com,  eelwjin@scut.edu.cn
}

\usepackage{bibentry}

\begin{document}

\maketitle

\begin{abstract}
Current multimodal large language models (MLLMs) face significant challenges in visual document understanding (VDU) tasks due to the high resolution, dense text, and complex layouts typical of document images. These characteristics demand a high level of detail perception ability from MLLMs. While increasing input resolution improves detail perception capability, it also leads to longer sequences of visual tokens, increasing computational costs and straining the models' ability to handle long contexts. To address these challenges, we introduce DocKylin, a document-centric MLLM that performs visual content slimming at both the pixel and token levels, thereby reducing token sequence length in VDU scenarios. We introduce an Adaptive Pixel Slimming (APS) preprocessing module to perform pixel-level slimming, increasing the proportion of informative pixels. Moreover, we propose a novel Dynamic Token Slimming (DTS) module to conduct token-level slimming, filtering essential tokens and removing others to adaptively create a more compact visual sequence. Experiments demonstrate DocKylin's promising performance across various VDU benchmarks and the effectiveness of each component.
\end{abstract}

\section{Introduction}
With the substantial growth in data and model sizes, as well as alignment with human preferences, current Large Language Models (LLMs) \cite{OpenAI_ChatGPT,touvron2023llama,vicuna,qwenlm,team2023internlm,yang2023baichuan} have demonstrated remarkable capabilities in reasoning, common sense understanding, expertise in specialized knowledge, zero/few-shot learning, and instruction following, shedding light on the potential for Artificial General Intelligence. Multimodal Large Language Models (MLLMs) aim to enhance these powerful LLMs by endowing them with multimodal capabilities, enabling the integration of diverse modalities beyond text alone \cite{yin2024survey,fu2024mme}.

Although there has been significant progress in the development of MLLMs, their performance in visual document understanding (VDU) tasks remains suboptimal \cite{liu2023hidden,zhang2024exploring}. A primary reason for this is their support for input resolutions of only up to $448\times448$, which is inadequate for high-resolution, fine-grained document images. Numerous efforts \cite{yu2024texthawk,liu2024textmonkey,li2024monkey,wei2025vary,bai2023qwen,ye2023ureader,lin2023sphinx,liu2024hrvda} have recently been made to increase the input resolution to address this limitation and achieved notable improvements. However, these efforts still face several challenges: \textbf{1)} Increasing the resolution also escalates the redundant regions within documents, which results in an increased number of redundant visual tokens, complicating the task for LLMs to identify the correct answers. \textbf{2)} Some methods \cite{li2024monkey,wei2025vary,liu2024hrvda,bai2023qwen} employ a fixed rectangular resolution, which can lead to text distortion given the typically large aspect ratios of document images. Moreover, scaling smaller images to larger ones results in inefficiencies. \textbf{3)} To manage the extended sequences of visual tokens resulting from high resolution, most methods \cite{li2024monkey,yu2024texthawk,liu2024textmonkey,bai2023qwen} employ a fixed compression ratio or extract a predetermined number of tokens. While this approach may be suitable for natural scene images, it is less effective for document scenarios where content density varies significantly. For example, images of scientific papers are information-dense, whereas images of identification cards are comparatively sparse.

To tackle these challenges, we propose DocKylin, a document-centric MLLM that incorporates several innovative designs: Firstly, we propose a parameter-free Adaptive Pixel Slimming (APS) preprocessing technique that utilizes gradient information to identify and eliminate redundant regions within document images, thereby reducing the proportion of redundant pixels and enhancing computational efficiency. Secondly, we adopt a more flexible image encoding strategy by capping the maximum number of pixels instead of fixing resolutions. This allows for variable resolutions and aspect ratios. Our lightweight Swin encoder \cite{liu2021swin} processes high-resolution document images holistically, ensuring feature coherence. Furthermore, we introduce a parameter-free Dynamic Token Slimming (DTS) method based on dual-center clustering. DTS efficiently filters informative tokens from a large set of visual tokens, resulting in reduced and dynamically sized visual sequences. Incorporating these innovations, we developed our DocKylin, which demonstrates superior performance across multiple VDU benchmarks. Extensive experiments demonstrate that DocKylin outperforms existing methods across various VDU benchmarks. Furthermore, both the proposed APS and DTS are parameter-free and modular, facilitating easy integration into existing MLLMs to achieve enhanced performances.

\section{Related Works}


\subsection{Text-centric MLLMs for VDU}
Visual document understanding (VDU) \cite{cui2021document,zhang2023docaligner,xu2020layoutlm, donut} centers on the automated processing, interpretation, classification, and extraction of information from document images with varied typesetting formats. This task is particularly challenging due to the diverse layouts \cite{cheng2023m6doc}, often poor image quality \cite{zhang2024docres,zhang2023appearance,zhang2022marior}, and complex structures within the documents.

Leveraging large language models (LLMs) trained on extensive datasets, emerging MLLMs have recently demonstrated significant potential in VDU tasks. However, their performance still falls short of expectations \cite{liu2023hidden,zhang2024exploring,tang2024mtvqa,shan2024mctbench}. Therefore, there has been a surge of recent work studying how to enhance the perception, recognition, and understanding abilities of MLLMs in such challenging scenarios. 

Some approaches \cite{tanaka2024instructdoc,kim2023visually,lu2024bounding,liao2024doclayllm} leverage additional OCR engines to recognize text in images and input it into MLLMs. Others \cite{li2024monkey,liu2024textmonkey,bai2023qwen,dong2024internlm,feng2023docpedia, ye2023ureader, wei2025vary, zhao2024tabpedia, zhao2024harmonizing} enhance the fine-grained perceptual capabilities of MLLMs by maintaining the high resolution of the input images. This includes strategies such as splitting the image into small pieces to accommodate the ViT's requirement for fixed and small-size inputs and utilizing visual encoders that are more suitable for text images. Authors of \cite{feng2023unidoc,wang2023towards} explore learning text grounding for MLLMs, while \cite{hu2024mplug,tang2024textsquare} investigate the use of more superior and larger scale instruction data to boost performance. Although these efforts have significantly improved the performance of MLLMs on text-centric tasks, there is still substantial room for improvement.

\subsection{Visual Token Compression in MLLMs}
Maintaining high-resolution inputs is a straightforward and effective method to enhance the performance of MLLMs on text images. However, increased resolutions inevitably lead to a longer visual sequence. To eliminate this problem, various methods are proposed for compressing the length of visual sequences in current MLLMs. One of the earlier and commonly used approaches \cite{li2023blip, wang2024pargo} involves utilizing a group of learnable query tokens to extract information in a cross-attention manner. Another method \cite{chen2023minigpt,yu2024texthawk} involves concatenating adjacent tokens along the channel dimension to reduce sequence length. Additionally, convolutional neural networks have also been proven effective for visual token compression \cite{cha2024honeybee,hu2024mplug}. However, all these methods either extract tokens of a fixed length or compress them at a fixed ratio, which is unsuited for document images, where the content density can vary significantly. A token sequence that is too short or a compression ratio that is too high can result in an inadequate representation. Conversely, longer token sequences or lower compression rates can reduce computational efficiency.

Recent studies \cite{cai2024matryoshka} have shown that the optimal length of visual token sequences varies for each sample. Consequently, some recent approaches have explored the adaptive compression of visual sequences. Most of these methods \cite{liu2024visual,shang2024llava,chen2025image,lin2024boosting} dynamically discard nonessential tokens using attention scores between visual tokens or between visual tokens and class tokens. However, considering the differences between text and natural objects with complete semantics, this paradigm may not be suitable for text images. For text-centric images, HRVDA \cite{liu2024hrvda} employs text position annotations to train a separate visual token filtering network, which requires introducing additional parameters and training procedures. Compared to the methods above, we explore alternative dynamic compression schemes specifically for document images, which include pixel-level redundancy removal and unsupervised clustering for filtering out nonessential tokens.

\section{Methodology}

\begin{figure}[t]
  \centering
  \includegraphics[width=3in]{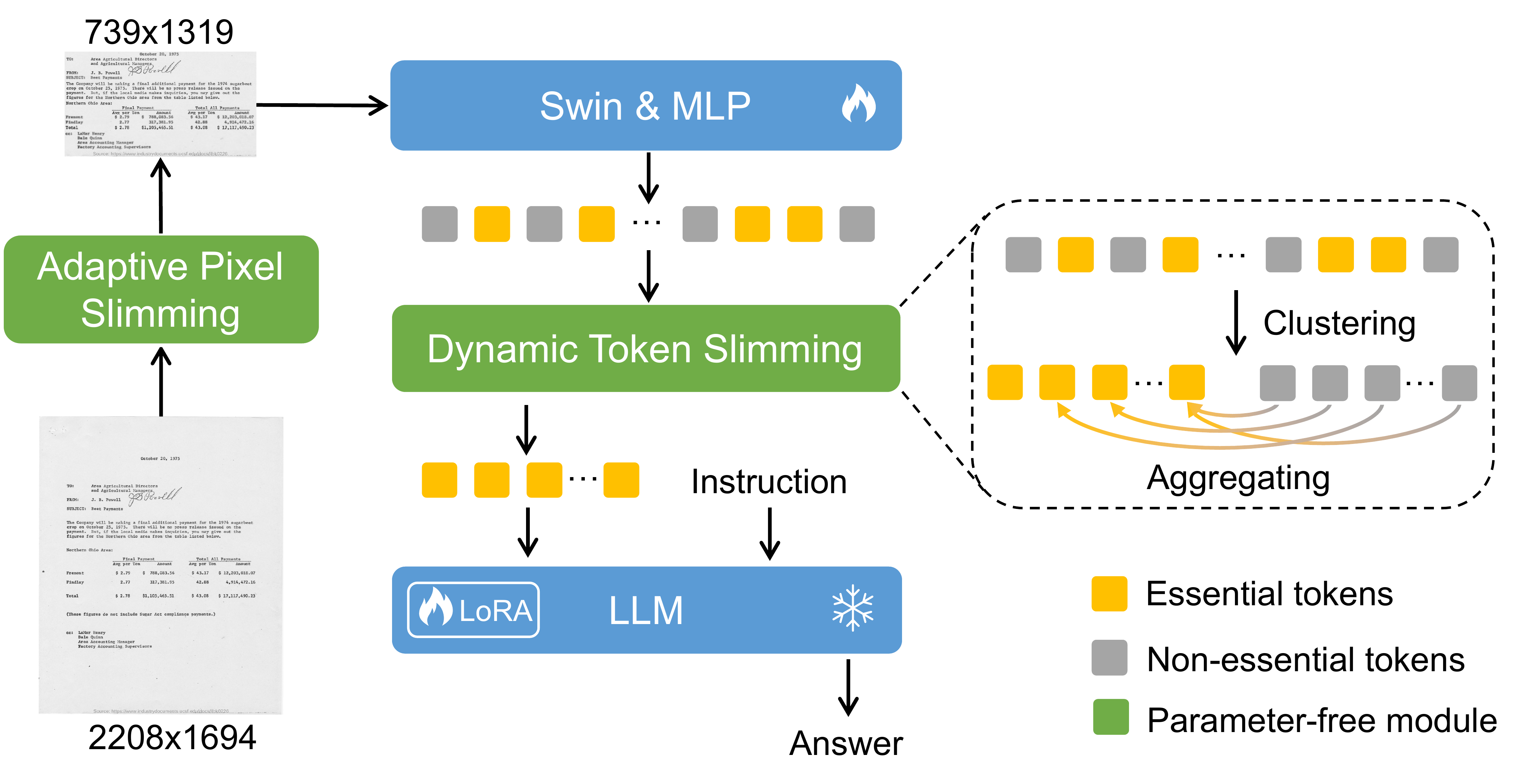}
  \caption{The overall architecture of our DocKylin model.}
  \label{fig:model}
\end{figure}

As illustrated in Fig.~\ref{fig:model}, the architecture of DocKylin incorporates an image encoder, an MLP layer that aligns the visual modality to the language modality, and a LLM. Building on this framework, DocKylin introduces the following enhancements: \textbf{1)} An Adaptive Pixel Slimming (APS) module preprocesses images before they are fed into the image encoder, which can remove redundant regions and reduce resolution while increasing the proportion of informative pixels. \textbf{2)} A more lightweight and flexible image encoder by capping the maximum number of pixels instead of fixing resolutions. \textbf{3)} A Dynamic Token Slimming (DTS) module handles the visual tokens produced by the visual encoder and MLP layer. DTS further removes nonessential tokens to enhance both performance and efficiency. 

In this paper, terms like `redundant' and `nonessential' refer to areas that can be masked without altering the document's overall meaning, allowing both humans and models to answer questions correctly. Examples include the background of the document or color blocks in bar charts.

\subsection{Adaptive Pixel Slimming}

Perceiving textual information in images typically requires maintaining a high input resolution, especially for text-rich document images. This poses significant challenges for current resource-intensive MLLMs. To address this issue, existing methods employ strategies such as dividing the images into patches for processing \cite{ye2023ureader}, employing visual encoders that support high resolution \cite{wei2025vary}, leveraging OCR results as additional inputs \cite{tanaka2024instructdoc,kim2023visually}, or applying compression transformations to the images beforehand \cite{feng2023docpedia}. 

\begin{figure*}[t]
  \centering
  \includegraphics[width=5in]{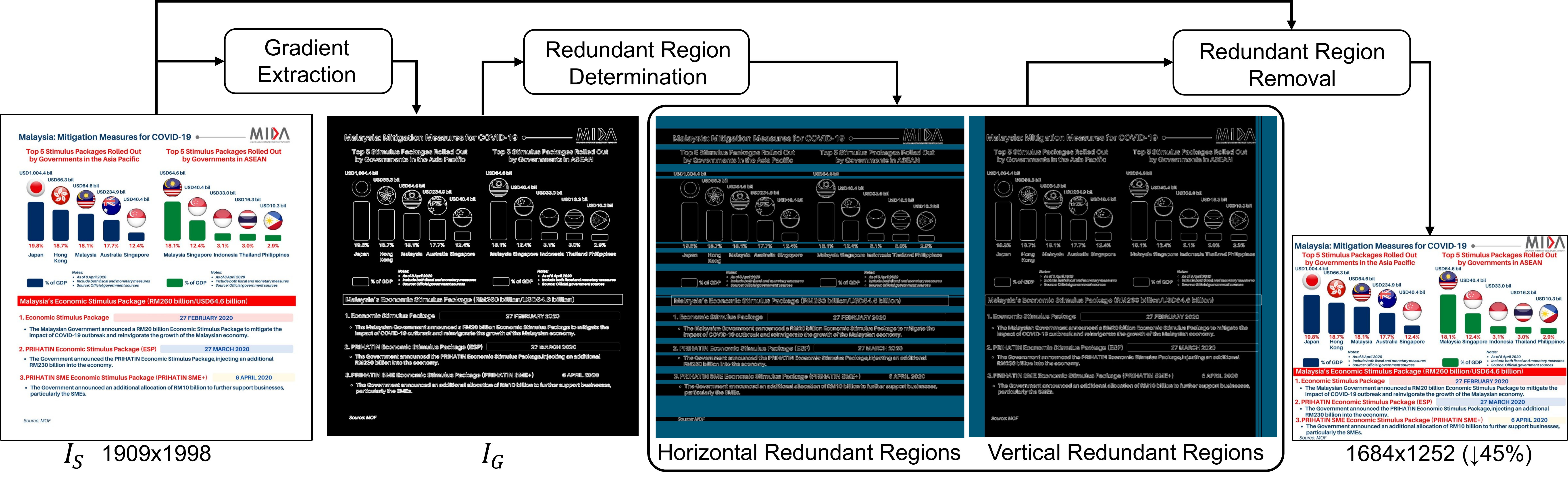}
  \caption{The proposed Adaptive Pixel Slimming module. It effectively reduces the resolution of document images by removing redundant regions.}
  \label{fig:Adaptive Pixel Slimming}
\end{figure*}

Here, orthogonal to the approaches mentioned above, we propose a novel approach called Adaptive Pixel Slimming (APS) to efficiently reduce the resolution of document images. APS is based on the observation that, as illustrated in Fig.~\ref{fig:Adaptive Pixel Slimming}, while document images usually exhibit high resolution, they commonly contain a significant proportion of redundant regions. These regions not only increase the computational burden on the visual encoder but may also pose challenges \cite{liu2024textmonkey,liu2024hrvda} to the reasoning capabilities of the LLMs by the increased redundant visual tokens. APS is thus designed to identify and remove these regions. Specifically, as shown in Fig.~\ref{fig:Adaptive Pixel Slimming}, during the gradient extraction stage, we utilize the Sobel operator to extract the gradient map $I_G$ from the original image $I_S$. Regions with low gradient values are considered smooth and devoid of significant information. Therefore, in the redundant region determination stage, we identify the redundant regions in the horizontal and vertical directions by locating areas with consistently low gradient values. For instance, to identify redundant regions in the image's horizontal direction: if the sum of gradient values for a specific row is below a predefined threshold, we classify that row as redundant. A sequence of contiguous redundant rows constitutes a redundant region in the horizontal direction. The vertical direction follows a similar procedure. The final image can be obtained by removing the identified redundant rows and columns. 

\subsection{Flexible and Lightweight Image Encoder}
Most existing MLLMs \cite{li2024monkey} utilize a fixed rectangular resolution, such as 224$\times$224 or 336$\times$336. This approach distorts the original aspect ratio of the images, leading to text deformation and inefficiencies when upscaling images smaller than the preset resolution. While some methods \cite{ye2023ureader,yu2024texthawk} allow for more flexible aspect ratios and resolutions, they often employ a patch-based processing strategy. Such a strategy may be suitable for natural scene images where the continuity and context of visual elements are less critical. However, it is less effective for document images that contain dense text \cite{liu2024textmonkey,huang2024mini}. Segmenting such images can disrupt the textual content, severing lines of text or splitting words and sentences across different patches. This discontinuity can impair the ability to accurately recognize and interpret textual information essential for effective document image understanding.

We employ a more flexible approach that can accommodate varying resolutions and aspect ratios without the need for image patching. Specifically, for an image obtained by APS with the size of $H\times W$, we set a maximum allowable number of pixels, MAX\_SIZE. If the product of H and W is less than MAX\_SIZE, the image will not be resized. Otherwise, a scaling factor $r=\mathrm{Int}(\sqrt{\frac{\mathrm{MAX\_SIZE}}{H\times W}})$ is calculated, and then the image is resized to $rH\times rW$. Here $\mathrm{Int}()$ denotes rounding down to the nearest integer to ensure that the resulting image does not exceed the maximum allowable pixels. One advantage of setting a maximum number of pixels rather than fixed dimensions is the ability to support more varied resolutions and preserve the original aspect ratio. Additionally, this manner avoids the loss of image details due to resolution reduction as much as possible without increasing computational resource consumption. For example, if we set the maximum pixel number MAX\_SIZE as 960$\times$960, we can accommodate an input resolution of 1280$\times$720 without alteration. In contrast, a fixed-dimension strategy would require resizing the image to 960$\times$960, resulting in a 25\% compression of the long side. Both strategies consume comparable computational resources since the number of visual tokens remains consistent.

Although ViT pre-trained with CLIP \cite{clip} demonstrates impressive generalization capabilities for natural scene images, they perform suboptimally on dense text images \cite{li2024enhancing, shan2024mctbench}. Furthermore, their self-attention mechanism, with its quadratic computational complexity scaling with image resolution, presents a significant challenge for processing high-resolution inputs. Consequently, we opt to use the more efficient Swin Transformer \cite{liu2021swin} as our visual encoder, which has been pre-trained on full-text recognition tasks \cite{donut}. It enables us to process an entire image as a single input.

\subsection{Dynamic Token Slimming}
Although the APS module has been effective in removing some redundant regions, it is limited to processing entire rows or columns and is ineffective for more complex scenarios. In contrast, our Dynamic Token Slimming (DTS) method offers a more flexible approach to eliminate redundant visual content at the token level. The design of the DTS is based on the premise that a well-trained visual encoder should be capable of effectively distinguishing between essential and nonessential regions within an image, with the features corresponding to different regions exhibiting sufficient discriminative properties. We propose to separate these two types of tokens directly through clustering and then aggregate nonessential tokens into essential ones. We next detail the proposed DTS, which comprises two main processes: Dual-center K-Means Clustering and Similarity Weighted Aggregation.

\textbf{Dual-center K-Means Clustering}. To distinguish tokens into essential and nonessential types without introducing additional parameters, we apply K-Means clustering with two centers to the output sequence from the visual encoder, as illustrated in Fig \ref{fig:model}. However, after obtaining the two clusters, it is unclear which cluster contains the essential tokens and which contains the non-essential tokens. Drawing on findings from \cite{liu2024textmonkey}, nonessential tokens typically lack uniqueness and are often similar to other tokens. Therefore, we identify potential nonessential tokens by evaluating the similarity of each visual token with all other ones. The cluster with fewer of these nonessential tokens is determined to be the one containing the essential tokens. For more implementation details, please refer to Algorithm \ref{al:clustering}.

\begin{algorithm}[t]\scriptsize
\caption{Dual-center K-Means Clustering}
\label{al:clustering}
\begin{algorithmic}[1]
\Require
\Statex visual tokens from Image Encoder: $V = \mathbb{R}^{L \times D}$
\Statex visual tokens from MLP Layer: $V' = \text{MLP}(V) \in \mathbb{R}^{L \times D'}$
\Statex CMX (calculate max similarity)
\Statex $R = 50$
\State $\text{cluster1, cluster2} = K\text{-means}(V, \text{center\_num}=2)$
\State max\_similarities = []
\State \textbf{for} $\text{token in } V'$:
\State \hspace{0.2cm} $\text{max\_similarities}.\text{append}(\text{CMX}(\text{token, other tokens in } V'))$
\State $\text{top-R\_similar\_tokens} = \text{select\_top\_similar\_tokens}(V', \text{max\_similarities}, R)$
\State num1 =\text{The number of common tokens between top-R\_similar\_tokens and cluster1}
\State num2 =\text{The number of common tokens between top-R\_similar\_tokens and cluster2}
\State \textbf{if} num1 \textgreater num2:
\State \hspace{0.3cm} essential\_tokens = cluster2
\State \hspace{0.3cm}nonessential\_tokens = cluster1
\State \textbf{else}:
\State \hspace{0.3cm} essential\_tokens = cluster1
\State \hspace{0.3cm} nonessential\_tokens = cluster2
\State \Return essential\_tokens, nonessential\_tokens
\end{algorithmic}
\end{algorithm}

\begin{figure}[t]
  \centering
  \includegraphics[width=3.1in]{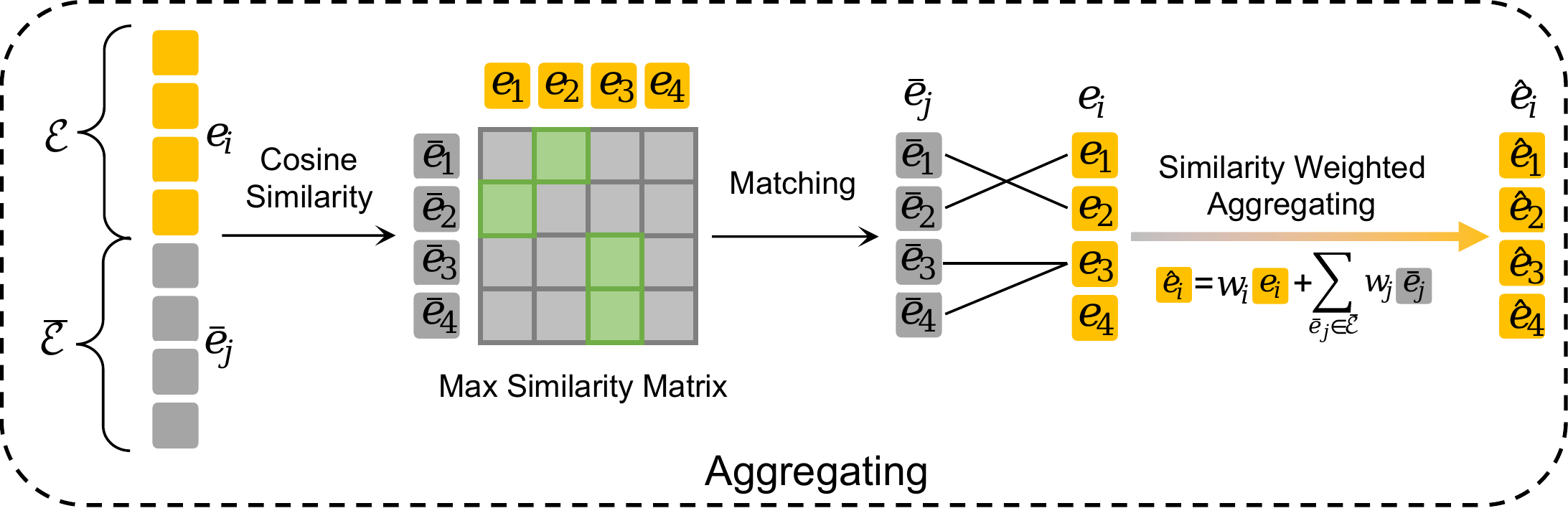}
  \caption{The proposed Similarity Weighted Aggregation module. It aggregates nonessential tokens into essential ones through similarity-weighted summation.}
  \label{fig:clustering}
\end{figure}

\begin{table*}[h]\scriptsize
\centering
\renewcommand{\arraystretch}{0.8}
\setlength{\tabcolsep}{1.3mm}{
\centering
\begin{tabular}{r|c|c|ccccccc}
\hline
Methods & Vis. Encoder (Params.) & Decoder & DocVQA$^{val}$ & InfoVQA$^{val}$ & ChartQA & FUNSD & SROIE & POIE \\ \hline
mPLUG-Owl \cite{ye2023mplug} & CLIP-L (0.3B) & LLaMA-7B & 7.4 & 20.0 & 7.9 &  0.2 & 0.1 & 0.3 \\
InstructBLIP \cite{dai2024instructblip} & CLIP-G (2B) & Vicuna-7B & 4.5 & 16.4 & 5.3 &  0.2 & 0.6 & 1.0 \\
LLaVA1.5 \cite{liu2024improved} & CLIP-L (0.3B) & LLaMA-7B & 8.5 & 14.7 & 9.3 & 0.2 & 1.7 & 2.5  \\
Qwen-VL \cite{bai2023qwen} & CLIP-G (2B) & Qwen-7B &  48.1 & 23.9 & 53.4 & 23.9 & 34.5 & 20.6 \\
Monkey \cite{li2024monkey} & CLIP-G (2B) & Qwen-7B & \underline{50.1} & 25.8 & \underline{54.0} & \underline{24.1} & \underline{41.9} & 19.9 \\
InternVL \cite{chen2023internvl} & IViT (6B) & Vicuna-7B & 28.7 & 23.6 & 45.6 & 6.5 & 26.4 & 25.9 \\ 
InternLM-XComposer2 \cite{dong2024internlm} & CLIP-L (0.3B) & InernLM2-7B & 39.7 & \underline{28.6} & 51.6 & 15.3 & 34.2 & \textbf{49.3} \\ \hline
\textbf{DocKylin (ours)} & Swin (0.07B) & Qwen-7B & \textbf{65.1} & \textbf{34.8} & \textbf{59.0} & \textbf{25.5} & \textbf{49.5} & \underline{36.1} \\ \hline
\end{tabular}}
\caption{Quantitative comparison of DocKylin with existing general-purpose MLLMs on document image benchmarks. Accuracy (\%) proposed in \cite{zhang2024exploring} is adopted as the metric for all benchmarks.}
\label{tab:sota}
\end{table*}

\begin{table*}[h]\scriptsize
\centering
\renewcommand{\arraystretch}{0.8}
\setlength{\tabcolsep}{1.3mm}{
\centering
\begin{tabular}{r|c|c|cccccc}
\hline
Methods & Visual Encoder (Params.) & Decoder & DocVQA$^{test}$ & InfoVQA$^{test}$ & ChartQA & VisualMRC & DeepForm  & WTQ  \\ \hline
Donut \cite{donut} & Swin (0.07B) & BERT(1-4 layers)& 67.5 & 11.6 & 41.8 & 93.9 & 61.6 &  18.8  \\
Pix2Struct \cite{lee2023pix2struct} & - & Transformers-1.3B & 76.6 & 40.0 & 58.6 & - & - & - \\
mPLUG-Doc \cite{ye2023mplug} & CLIP-L (0.3B) & LLaMA-7B & 62.2 & 38.2 & 57.4 & 188.8 & 42.6 & 26.9 \\ 
UReader \cite{ye2023ureader} & CLIP-L (0.3B) & LLaMA-7B & 65.4 & 42.2 & 65.7 & 221.7 & 49.5 & 29.4 \\
DocPeida \cite{feng2023docpedia} & Swin (0.1B) & Vicuna-7B & 47.1 & 15.2 & 46.9 & - & - & -\\ 
HRVDA \cite{liu2024hrvda} & Swin & LLaMA2-7B & 72.1 & 43.5 &  \textbf{67.6} & 211.5 & 63.2 & 31.2 \\
Vary \cite{wei2025vary} & \makecell{ViTDet-B \& CLIP-L \\(0.4B)} & Vicuna-7B & 76.3 & - & 66.1 & - &- &-\\
TextMonkey \cite{liu2024textmonkey} & CLIP-G (2B) & Qwen-7B & 71.5 & - & 65.5 & - & 61.6 & 30.6 \\
TextHawk \cite{yu2024texthawk}  & SigLIP-SO (0.4B) & InternLM-7B & \underline{76.4} & \textbf{50.6} & 66.6 & - & - & \textbf{34.7} \\
LayoutLLM \cite{luo2024layoutllm}& LayoutLMv3-L (0.3B) & LLaMA2-7B & 74.3 & - & - & - & -&-\\
 \hline
\textbf{DocKylin (ours)} & Swin (0.07B) & Qwen-7B & \textbf{77.3} & \underline{46.6} & \underline{66.8} & \textbf{319.9} & \textbf{64.2} & \underline{32.4}  \\ \hline
\end{tabular}}
\caption{Quantitative comparison of DocKylin with existing text-centric MLLMs on document-oriented benchmarks. The metrics used for each benchmark are consistent with those proposed in the original papers.}
\label{tab:sota2}
\end{table*}

\textbf{Similarity Weighted Aggregation.} After obtaining essential and nonessential tokens, how to handle the nonessential tokens becomes a crucial issue. Inspired by \cite{kong2022spvit,liang2021evit,wei2023joint}, to mitigate potential clustering errors and prevent information loss, we do not discard the nonessential tokens outright. Instead, we aggregate them into the essential tokens to minimize the potential loss of effective information while enhancing the model's processing efficiency. As illustrated in Fig. \ref{fig:clustering}, for each nonessential token from the clustering step, we find the most similar essential token. Each essential token may correspond to zero or multiple nonessential tokens. If an essential token has corresponding nonessential tokens, we perform a similarity-weighted summation of these tokens to obtain an aggregated token.

Specifically, after determining the set of essential tokens $\mathcal{E}=\{\boldsymbol{e}_1, \boldsymbol{e}_2, \boldsymbol{e}_3, \ldots, \boldsymbol{e}_N\}$ and the set of nonessential tokens $\mathcal{\overline {E}}=\{\overline {\boldsymbol{e}}_1, \overline {\boldsymbol{e}}_2, \overline {\boldsymbol{e}}_3, \ldots, \overline {\boldsymbol{e}}_{\overline {N}}\}$, we aggregate nonessential tokens $\overline {\boldsymbol{e}}_{j} $ into essential tokens $\boldsymbol{\boldsymbol{e}}_i $ by similarity weighted summation:
\begin{equation}\footnotesize
\hat{\boldsymbol{e}}_i=w_i\boldsymbol{e} _i+\sum_{ \overline {\boldsymbol{e}} _j\in \mathcal{\overline {E}}} w_j \overline {\boldsymbol{e}}_j,
\end{equation}
where $w_i$ and $w_j$ are the weight for the $i$-th essential and $j$-th nonessential token, and $\hat{\boldsymbol{e}}_i$ is the aggregated token. We use the similarity between $e_i$ and $\overline {e}_j$ to determine $w_i $ and $w_j$. Specifically, we first define the similarity matrix between essential tokens and nonessential tokens:
\begin{equation}\footnotesize
c_{i,j}=\frac{\boldsymbol{e}_i^T\overline {\boldsymbol{e}}_j}{\|\boldsymbol{e}_i\|\|\overline {\boldsymbol{e}}_j\|}.
\end{equation}
Next, we obtain a mask matrix to ensure that each nonessential token is aggregated only with the most similar essential token:
\begin{equation}\footnotesize
m_{i,j}=\begin{cases}1,&i=\mathop{\arg\max}\limits_{\overline {\boldsymbol{e}}_j\in \mathcal{\overline {E}}}c_{i,j},\\0,&\mathrm{otherwise},\end{cases}.
\end{equation}
Finally, we can obtain aggregating weight $w_j$ and $w_i$ by:
\begin{equation}\footnotesize
w_j=\frac{\exp(c_{i,j})m_{i,j}}{\sum_{\overline {\boldsymbol{e}}_j\in \mathcal{\overline {E}}}\exp(c_{i,j})m_{i,j}+\mathrm{e}}
\end{equation}
and
\begin{equation}\footnotesize
w_i=\frac{e}{\sum_{\overline {\boldsymbol{e}}_j\in \mathcal{\overline {E}}}\exp(c_{i,j})m_{i,j}+\mathrm{e}}.
\end{equation}

\begin{figure}[t]
  \centering
  \includegraphics[width=3.3in]{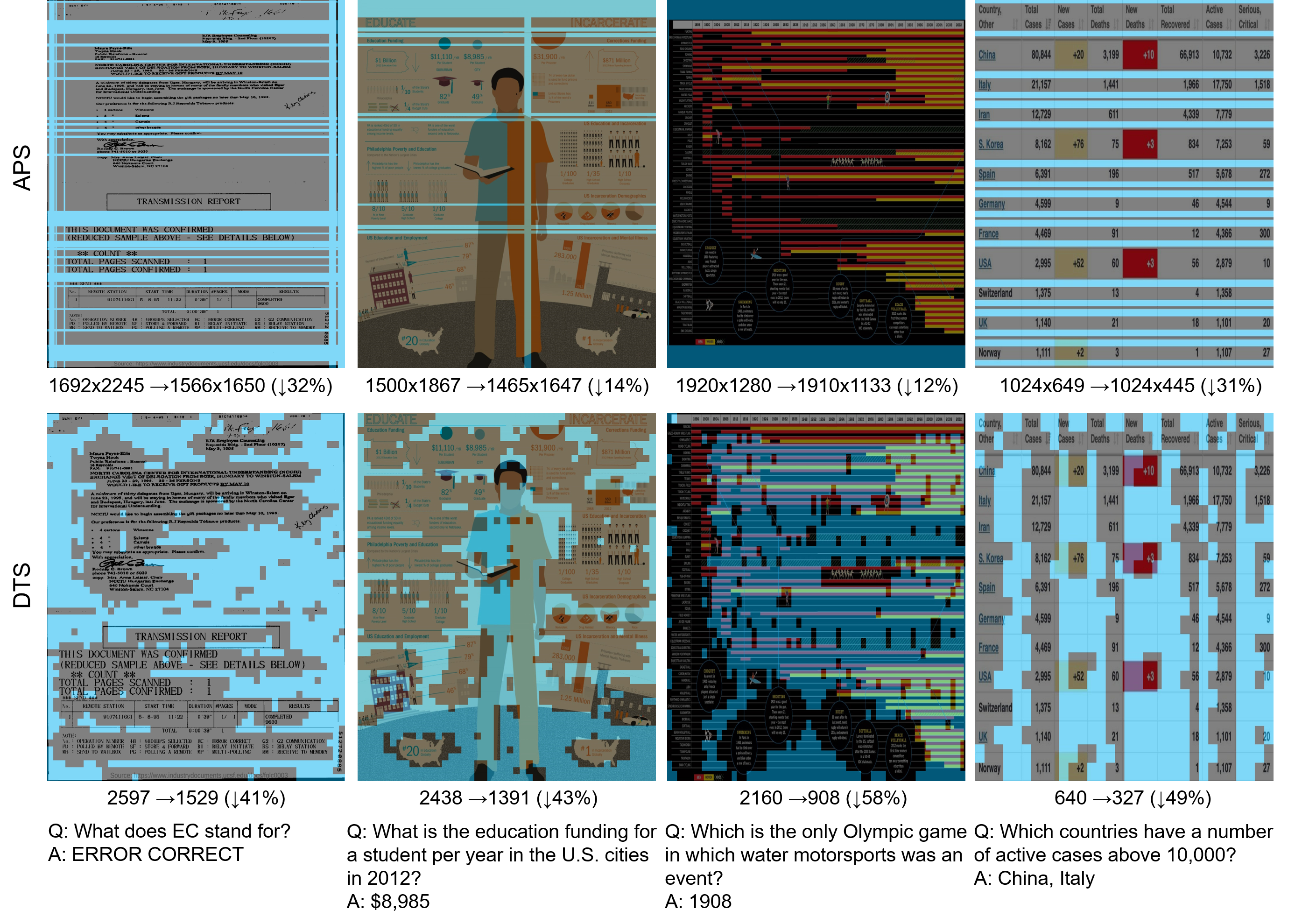}
  \caption{Visualization from DocKylin. The redundant pixels and nonessential tokens identified by Adaptive Pixel Slimming and Dynamic Token Slimming are highlighted in light blue. Zoom in for best view.}
  \label{fig:visualize}
\end{figure}

\begin{table*}[t]\scriptsize
\centering
\renewcommand{\arraystretch}{0.8}
\begin{tabular}{ccccccccc}
\hline
\multirow{2}{*}{Model} & \multirow{2}{*}{Compression} & \multirow{2}{*}{Adaptive} & \multicolumn{2}{c}{DocVQA$^{val}$} & \multicolumn{2}{c}{InfoVQA$^{val}$} & \multicolumn{2}{c}{SROIE} \\ \cline{4-9} 
 & & & Avg. len & Acc $\uparrow$ & Avg. len & Acc $\uparrow$ & Avg. len & Acc $\uparrow$\\ \hline
\multirow{11}{*}{LLaVA1.5 \cite{liu2024improved}} & - & - & 576 & 8.5 & 576 & 14.7 & 576 & 1.7\\
 & Random & \XSolidBrush  & 300 & 5.4 & 300 & 14.3 & 300 & 0.5 \\
 & Random & \XSolidBrush  & 200 & 4.2 & 200 & 14.2 & 200 & 0\\
& MaxPool1D & \XSolidBrush  & 288 & 6.7 & 288 & 13.8 & 288 & 0.5\\
& AvgPool1D & \XSolidBrush  & 288 & 7.3 & 288 & 14.5 & 288 & 1.1\\
 & FastV (K=2, R=50\%) \cite{chen2025image} & \XSolidBrush  & 306 & 8.4 & 306 & 14.4 & 306 & 2.0\\
 & PruMerge \cite{shang2024llava} & \Checkmark & \textbf{123} & 5.3 & \textbf{123} & 13.8 & \textbf{121} & 0.4 \\
 & PruMerge+ \cite{shang2024llava} & \Checkmark  & 232 & 5.2 &233&14.0&229&0.3\\ \cline{2-9}
 & APS+DTS w/o Aggregation (ours) & \Checkmark  & 298 & \textbf{9.9} & 270 & 14.2 & 296 & \textbf{3.4}\\
 & APS+DTS (ours) & \Checkmark  & 298 & \textbf{9.9} & 270 & \textbf{14.9} & 296 & 3.3\\
 \hline
\end{tabular}
\caption{Comparisons with current SOTA parameter-free visual token compression methods. Accuracy proposed in \cite{zhang2024exploring} is adopted as the metric for all benchmarks. `DTS w/o Aggregation' refers to skipping the Similarity Weighted Aggregation process mentioned earlier and directly discarding the nonessential tokens.}
\label{tab:toy}
\end{table*}

\section{Experiments}

\subsection{Implementation Details}
Our model underwent two training stages: pre-training and instruction tuning. All the training data are sourced from open-access datasets. Due to page limitations, specific datasets used are detailed in the supplementary materials. During the pre-training stage, we trained only the visual encoder and the MLP, keeping the LLM parameters fixed. In this phase, the model learns to perform full-text recognition on document images and convert charts and tables into markdown formats \cite{hu2024mplug}. The goal is to enable the visual encoder to comprehend document content and align the visual features with the language space through the MLP. 

During the pre-training phase, the model was trained for 450,000 iterations with a batch size of 8. The learning rate decayed from 1e-4 to 1e-5 using cosine annealing, which took approximately 40 A800 GPU days. In the instruction tuning phase, the model underwent 300,000 iterations with the same batch size and learning rate as in the pre-training phase. This phase further consumed approximately 24 A800 GPU days. Our Adaptive Pixel Slimming method is employed in both phases, while Dynamic Token Slimming (DTS) is used only during the instruction tuning phase. The maximum allowed number of pixels, MAX\_SIZE, is set to 1728$\times$1728. We initialize our visual encoder with Donut-Swin (0.07B) \cite{donut}, pre-trained on the reading texts task, and use Qwen-7B-Chat \cite{qwenlm} as our LLM.

\subsection{Comparisons with Prior Arts}

We first compare our model with several existing general-purpose multimodal large language models (MLLMs). As these models sometimes tend to provide detailed responses, following \cite{zhang2024exploring}, we adopt accuracy as our metric: a response is considered correct if it contains the complete answer. The results, as shown in Table \ref{tab:sota}, demonstrate that DocKylin achieves a significant advantage over the competing models. This highlights that VDU tasks remain challenging for current general-purpose MLLMs. Note that descriptions of all these benchmarks are included in the supplementary materials.

We also compare our model with several existing text-centric MLLMs. The results, presented in Table \ref{tab:sota2}, use the same metrics as those employed in the original benchmarks. Our method demonstrates advantages over these text-centric MLLMs, achieving leading performance across multiple benchmarks. Notably, our visual encoder is more lightweight than those used by the compared methods. 

Some visual results from DocKylin on these benchmarks are presented in Fig. \ref{fig:visualize}. The first row shows the results of redundant regions detected by APS, with corresponding text explaining the resulting reduction in resolution. The second row displays the nonessential tokens removed by DTS, with accompanying text illustrating its effect on compressing the visual sequence. These results demonstrate their effectiveness in accurately retaining important regions while significantly reducing the resolution and the length of the visual sequence.

\begin{table}[t]\scriptsize
\centering
\renewcommand{\arraystretch}{0.8}
\setlength{\tabcolsep}{0.8mm}{
\begin{tabular}{cc|c|c|c|cccc}
\hline
APS & DTS & Token Len. & Train & Infer. & DocVQA$^{val}$ & ChartQA  & InfoVQA & POIE\\ \hline
\XSolidBrush         & \XSolidBrush                    & 2770 & 1    & 1  & 63.1   &     55.8   & \textbf{34.8} & 34.5 \\
\Checkmark           & \XSolidBrush                    & 2198 & 0.97    & 0.81 & 64.5  & 58.5  & 34.7 & 34.9 \\
\Checkmark           & \Checkmark                      & \textbf{1250}                   & \textbf{0.86} & \textbf{0.61} & \textbf{65.1}  & \textbf{59.0} & \textbf{34.8} & \textbf{36.1} \\ \hline
\end{tabular}}
\caption{The training and inference time are normalized to the results in the first row. The token length and inference time are obtained based on the DocVQA$^{val}$ dataset. Accuracy from \cite{zhang2024exploring} is adopted as the metric.}
\label{tab:ablation}
\end{table}

To further demonstrate the superiority of our method, we compared it with other parameter-free visual token compression methods, using LLaVA1.5 \cite{liu2024improved} as our baseline model (without retraining). The results in Table \ref{tab:toy} indicate that our APS and DTS achieve performance improvements with shorter sequence lengths than the baseline model, significantly outperforming existing SOTA compression methods. We also found that incorporating weighted aggregation provided a slight performance boost. After model training, we expect enhanced benefits \cite{shang2024llava,chen2025image}, as aggregation introduces an unfamiliar pattern to the model. Theoretically, it offers a higher performance ceiling by retaining more original information.

\begin{table*}[]\scriptsize
\centering
\renewcommand{\arraystretch}{0.8}
\begin{tabular}{ccccccccccccc}
\hline
\multirow{2}{*}{Strategy} & \multicolumn{2}{c}{Scanned Document} & \multicolumn{2}{c}{Infographics} & \multicolumn{2}{c}{Chart} & \multicolumn{2}{c}{Table} & \multicolumn{2}{c}{Receipt} & \multicolumn{2}{c}{Ingredient List} \\ \cline{2-13} 
 & Mask Ratio & Acc.$\uparrow$ & Mask Ratio & Acc.$\uparrow$ & Mask Ratio & Acc.$\uparrow$ & Mask Ratio & Acc.$\uparrow$ & Mask Ratio & Acc.$\uparrow$ & Mask Ratio & Acc.$\uparrow$ \\ \hline
None & - & 55.2 & - & 26.7 & - & 73.5 & - & 34.8 & - & 44.0 & - & 25.9 \\ \hline
APS-R-mask & 63.4 & 0.2 & 83.3 & 14.4 & 68.5 & 3.8 & 96.0 & 4.8 & 36.1 & 0.1 & 99.4 & 0.1 \\
DTS-R-mask & 51.8 & 0.2 & 64.8 & 14.6 & 57.6 & 4.4 & 61.9 & 0.6  & 55.9 & 0.1 & 66.0 & 0.5\\ \hline
Top-30-mask & 30 & 50.4 & 30 & 24.0 & 30 & 74.3 & 30 & 25.8 & 30 & 37.1 & 30 & 19.7 \\ 
Top-40-mask & 40 & 45.3 & 40 & 22.3 & 40 & 71.0 & 40 & 23.1 & 40 & 32.3 & 40 & 15.4 \\ 
\hline
APS-mask & 36.6 & 54.7 & 16.7 & 26.4 & 31.5 & 74.8 & 4.0 & 34.6 & 63.9 & 44.0 & 0.6 & 26.1 \\
DTS-mask & 48.2 & 54.9 & 35.2 & 26.3 & 42.4 & 73.8 & 38.1 & 31.5 & 44.1 & 43.2 & 34.0 & 25.1 \\
\hline
\end{tabular}
\caption{The Scanned Document, Infographics, Chart, Table, Webpage, Receipt, and Ingredient List are sourced from the DocVQA$^{val}$, InfoVQA$^{val}$, ChartQA (augmented), WTQ, visualMRC, SROIE and POIE datasets, respectively. All data, except for the Webpage which uses the CIDEr metric, are evaluated using the accuracy metric as proposed in \cite{zhang2024exploring}.}
\label{tab:mask}
\end{table*}

\subsection{Ablation Studies}
\textbf{The Effectiveness of APS and DTS on DocKylin}. We validate the effectiveness of APS and DTS on DocKylin and present the results in Table \ref{tab:ablation}. Note that, settings without DTS in Table \ref{tab:ablation} refer to retraining a new model from scratch without employing DTS in the instruction tuning phase, rather than simply removing DTS from the already trained DocKylin model. The results demonstrate that both APS and DTS effectively reduce token sequence length and enhance computational efficiency, while also improving performance in most cases. This aligns with findings from previous studies \cite{liu2024textmonkey}, which suggest that excessively long sequences due to high resolution may degrade performance. This degradation likely occurs because the visual sequences contain too many redundant tokens, placing greater demands on the model's ability to handle long contexts and making the learning more challenging. Furthermore, the APS and DTS methods can also be used during the training phase to accelerate training. Specifically, applying the DTS method during the instruction tuning phase can reduce training time by approximately 15\%.

\begin{table*}[t]\scriptsize
\centering
\renewcommand{\arraystretch}{0.8}
\setlength{\tabcolsep}{1mm}{
\begin{tabular}{c|c|ccccc}
\hline
Methods & Resolution & DocVQA$^{val}$ & InfoVQA$^{val}$ & SROIE & FUNSD  \\ \hline
LLaVA1.5 \cite{liu2024improved}  & \multirow{2}{*}{336$\times$336} &  8.5 & 14.7 & 1.7 & 0.2  \\
LLaVA1.5 \cite{liu2024improved} + APS &  &  10.7 (\textbf{+27.4\%})
 & 14.7 (\textbf{+0\%}) & 3.7 (\textbf{+118\%}) & 0.92 (\textbf{+360\%})  \\ \hline
Qwen-VL \cite{bai2023qwen} & \multirow{2}{*}{448$\times$448} &  48.1 & 23.9 & 34.5 & 20.6  \\
Qwen-VL \cite{bai2023qwen} + APS &  &  51.2 (\textbf{+6.4\%})
 & 24.7 (\textbf{+4.1\%}) & 40.0 (\textbf{+15.9\%}) & 24.3 (\textbf{+17.9\%})  \\ \hline
 Monkey \cite{li2024monkey} & \multirow{2}{*}{896$\times$896} &  50.1 & 25.8 & 41.9 & 24.1  \\
Monkey \cite{li2024monkey} + APS & & 56.3 (\textbf{+12.4\%})
 & 27.5 (\textbf{+6.6\%}) & 47.0 (\textbf{+12.2\%}) & 27.3 (\textbf{+13.3\%})  \\  \hline
  InternVL2 \cite{chen2024far} & \multirow{3}{*}{448$\times$448$\times$(1$ \sim $12)} &  76.2&49.5&54.7&41.7  \\
InternVL2 \cite{chen2024far} + APS & & 76.1&48.2&54.2&40.6  \\
InternVL2 \cite{chen2024far} + APS + Resize & & 77.3 (\textbf{+1.4\%})&49.4 (\textbf{-0.2\%})&55.2 (\textbf{+0.9\%})&43.4 (\textbf{+4.1\%})  \\ 
 \hline
\end{tabular}}
\caption{The enhancement brought to other MLLMs by Adaptive Pixel Slimming. 
\textbf{Relative improvements} are highlighted in bold. Accuracy proposed in \cite{zhang2024exploring} is adopted as the metric for all benchmarks.}
\label{tab:Adaptive Pixel Slimming}
\end{table*}

\begin{figure}[t]
  \centering
  \includegraphics[width=2.1in]{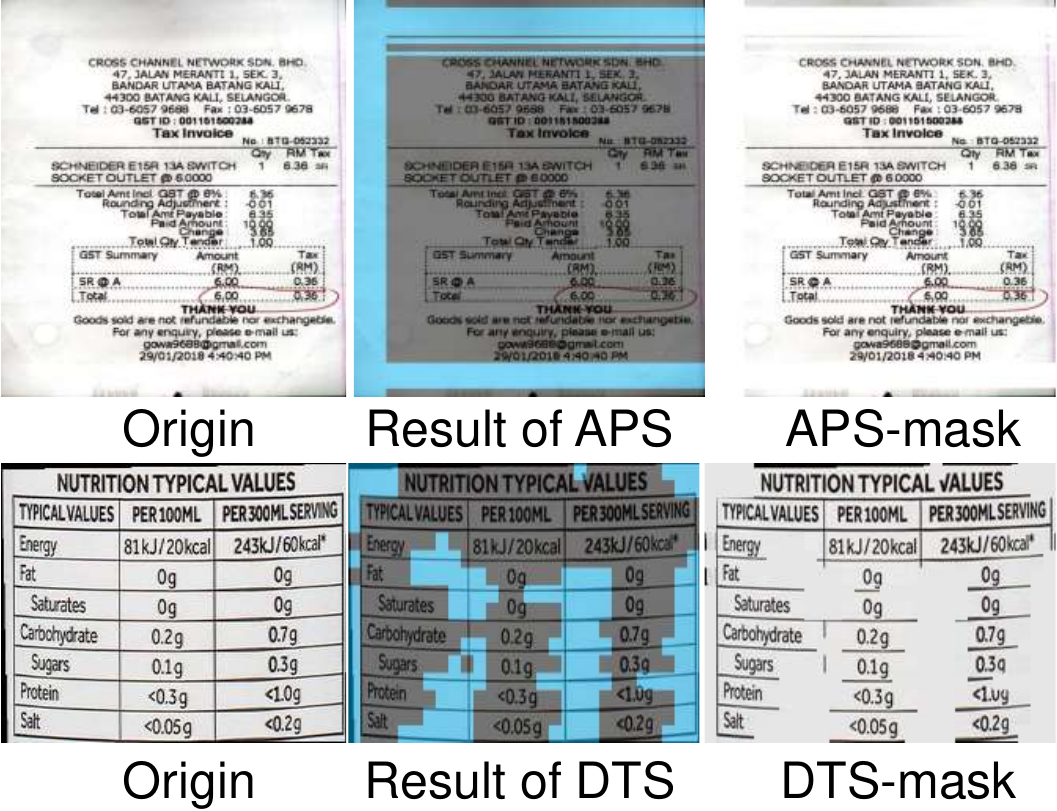}
  \caption{The results of APS-mask and DTS-mask. The identified redundant regions are highlighted in light blue. To minimize any additional effects caused by masking, the values in the masked regions are set to the average value of the pixels in the current region.}
  \label{fig:mask}
\end{figure}

\textbf{The accuracy of Identified Redundant Regions}. We conduct experiments to further explore: \textbf{1)} How accurately do APS and DTS identify redundant regions? \textbf{2)} Do APS and DTS demonstrate consistent performance across different data types? Specifically, we apply APS and DTS to identify redundant regions in different document images, and then mask these regions instead of removing them (as shown in Fig. \ref{fig:mask}). The masked images are used as inputs to an existing MLLM to assess any performance degradation. This approach allows us to isolate the accuracy of APS and DTS from the performance improvements brought by redundancy removal.

Table \ref{tab:mask} shows that APS effectively reduces resolution across all image types with minimal performance impact. However, the reduction effects on the table and ingredient list are less pronounced, likely due to the lines spanning the entire image, making APS difficult to identify redundant rows or columns. DTS also effectively reduces resolution across all image types and typically achieves greater resolution reduction than APS, as it can identify local areas beyond entire rows and columns. While DTS may cause noticeable performance drops in the table datasets, our visual analysis suggests this is due to new patterns introduced by masking table lines, not the masking of critical text content. Practically, such potential performance degradation from DTS can be largely mitigated: \textbf{1)} DTS retains information by similarity-weighted aggregation approach rather than completely discarding redundant tokens; and \textbf{2)} during image encoding, information from masked areas is already integrated into neighboring tokens. 

We conduct additional experiments by reversing the APS-mask and DTS-mask results by masking the non-redundant regions identified by APS and DTS, denoted as APS-R-mask and DTS-R-mask. The results indicate a significant performance drop, further demonstrating the accuracy of APS and DTS in identifying redundant regions. 

We further compare our approach with methods that directly select redundant tokens based on similarity priors (e.g., TextMonkey and PruMerge adopt this approach). Specifically, in Table 5, Top-30-mask and Top-40-mask represent masking the top 30\% and 40\% of tokens with the highest similarity to other tokens, respectively, as nonessential tokens. Results show that DTS outperforms the top K strategy, even at higher reduction rates.

\textbf{Potentials for Further Application and Research}. Since APS is a parameter-free method that requires no additional training or modification of model architecture, it can be easily applied to other MLLMs to enhance their performance on high-resolution document images. To further demonstrate its effectiveness, we apply APS to various existing MLLMs (employing it before the image processing module of these models). As shown in Table~\ref{tab:Adaptive Pixel Slimming}, APS consistently improves the performance of MLLMs that support lower resolutions (LLaVA1.5, Qwen-VL and Monkey). However, directly applying APS to high-resolution models like InternVL2 results in a slight performance decline, likely due to a mismatch between the dense text produced by APS and the high-resolution training setup of InternVL2. Adjusting the APS-processed images back to their original resolution (InternVL2+APS+Resize) can achieve a performance gain of up to 4.1\%. This suggests that APS alone is most effective for low-resolution models, while APS+Resize is more suitable for high-resolution models.

\begin{figure}[t]
  \centering
  \includegraphics[width=3.in]{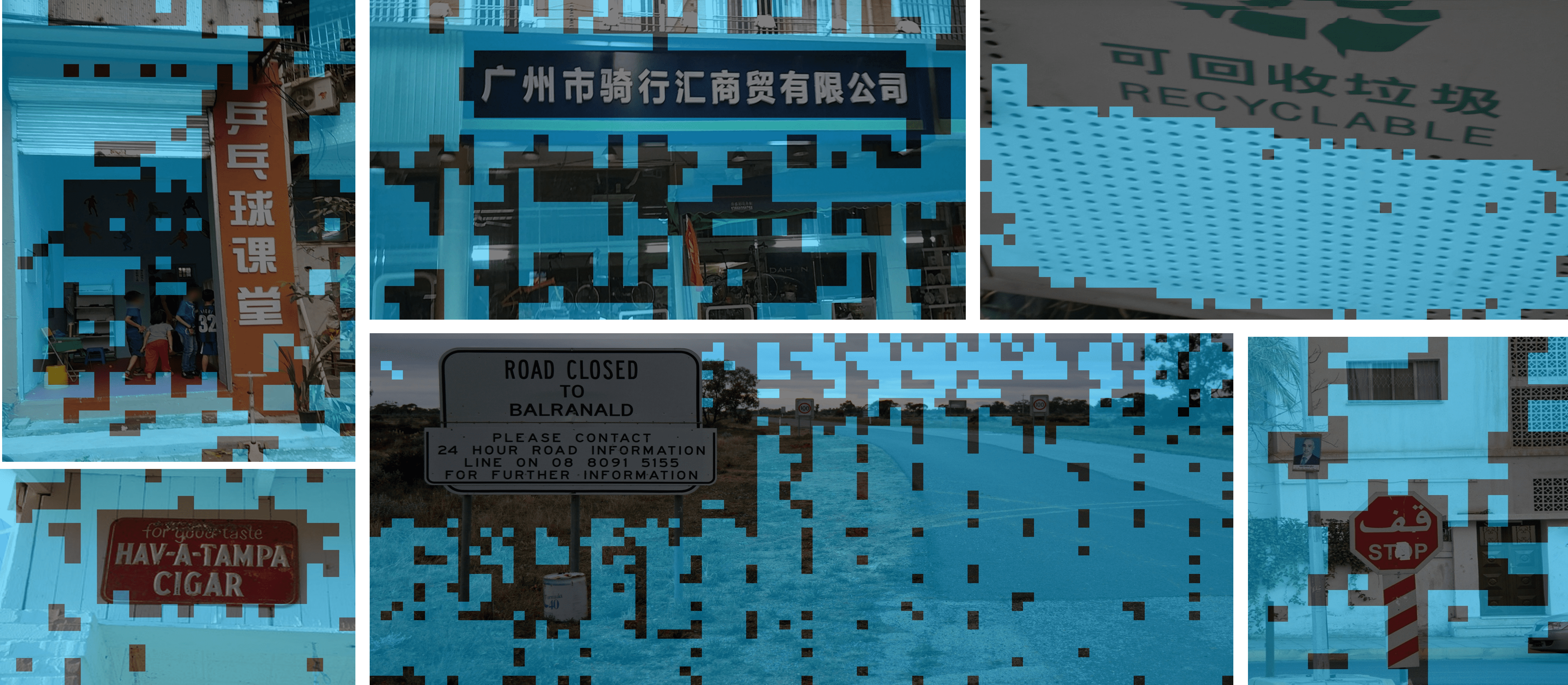}
  \caption{Effects of applying DTS to natural scene images.}
  \label{fig:scene_text}
\end{figure}

The redundant regions in document images are often simple, such as plain backgrounds, distinguishing these regions is relatively easy. This raises the question of whether DTS's effectiveness is due to this simplicity. Although this paper primarily explores VDU tasks, we are also interested in the generalizability of DTS. Therefore, we conducted some experiments on scene text images \cite{tang2022few,liu2023spts,zhang2020sahan}. Specifically, we retrained a scene text image full-text recognition model using some open-source scene text data, performing only the first pre-training phase. We then applied DTS to the trained model and tested it on the TextVQA dataset. Some visual results are shown in Fig. \ref{fig:scene_text}. It can be seen that DTS is effective in distinguishing nonessential regions even in the complex backgrounds of natural scene images.

\section{Conclusions}
To address the issues of high input resolution and excessively long token sequences in current MLLMs when counters with VDU tasks, we propose the DocKylin model. It includes an Adaptive Pixel Slimming (APS) preprocessing module that removes redundant regions from document images, effectively reducing their resolution. The model employs a lightweight and flexible image encoding approach to adapt to the varying resolutions and aspect ratios of document image scenarios. Moreover, we propose a token compression module, Dynamic Token Slimming (DTS), which clusters and selects essential tokens from a long visual sequence and employs aggregation to avoid potential information loss. Experiments demonstrate that DocKylin achieves leading performance across multiple VDU benchmarks. Extensive experiments demonstrate that APS and DTS effectively reduce the length of visual tokens, enhancing both computational efficiency and model performance. The parameter-free nature of our APS and DTS modules not only enhances efficiency but also facilitates easy integration into existing multimodal large language models, and our experiments indicate their potential for broader applications and scenarios. This feature makes our approach particularly valuable for improving document understanding capabilities in a wide range of applications.

\section{Acknowledgments}
This research is supported in part by the National Natural Science Foundation of China (Grant No.: 62441604, 62476093).

\bibliography{aaai25}

\clearpage
\appendix
\section{Appendix}
\subsection{Training Data}

\begin{table}[h]\scriptsize
\caption{Datasets used in pre-training and instruction tuning phase.}
\label{tab:dataset}
\centering
\begin{tabular}{c|p{4.5cm}}
\hline
Stage        & Datasets                                      \\ \hline
\multirow{16}{*}{Pre-training} & CASIA-HWDB \cite{liu2011casia}, CCpdf \cite{turski2023ccpdf}, ChartQA \cite{masry2022chartqa}, DeepForm \cite{svetlichnaya2020deepform}, DocVQA \cite{mathew2021docvqa}, DVQA \cite{kafle2018dvqa}, FigureQA \cite{kahou2017figureqa}, InfoVQA \cite{mathew2022infographicvqa}, KLC \cite{stanislawek2021kleister}, OCRVQA \cite{mishra2019ocr}, PlotQA \cite{methani2020plotqa}, PubTabNet \cite{zhong2020image}, RVL-CDIP \cite{harley2015evaluation}, SCUT-HCCDoc \cite{zhang2020scut}, SynthDog \cite{donut}, TAT-QA \cite{zhu2021tat}, VisualMRC \cite{tanaka2021visualmrc}, WildReceipt \cite{sun2021spatial} \\ \hline
\multirowcell{10}{Instruction\\ tuning} & ChartQA \cite{masry2022chartqa}, DeepForm \cite{svetlichnaya2020deepform}, DocILE \cite{vsimsa2023docile}, DocVQA \cite{mathew2021docvqa}, DVQA \cite{kafle2018dvqa}, InfoVQA \cite{mathew2022infographicvqa}, InsuranceQA \cite{insuranceqa}, KLC \cite{stanislawek2021kleister}, OCR-TROSD\cite{readingorder}, TAT-QA \cite{zhu2021tat}, VisualMRC \cite{tanaka2021visualmrc}, WildReceipt \cite{sun2021spatial}, WTQ \cite{wtq} \\ \hline
\end{tabular}
\end{table}
The training data for the two stages are summarized in Table \ref{tab:dataset}.

\begin{table}[t]\scriptsize
\caption{The enhancement brought to other MLLMs by Adaptive Pixel Slimming. 
\textbf{Relative improvements} are highlighted in bold. Accuracy proposed in \cite{zhang2024exploring} is adopted as the metric for all benchmarks.}
\label{tab:Adaptive Pixel Slimming supp}
\centering
\renewcommand{\arraystretch}{0.8}
\setlength{\tabcolsep}{1mm}{
\begin{tabular}{c|c|c}
\hline
Methods & Resolution & TextVQA \\ \hline
LLaVA1.5 \cite{liu2024improved}  & \multirow{2}{*}{336$\times$336} & 38.7  \\
LLaVA1.5 \cite{liu2024improved} + APS &  & 40.2 (\textbf{+4.1\%}) \\ \hline
Qwen-VL \cite{bai2023qwen} & \multirow{2}{*}{448$\times$448} & 62.1  \\
Qwen-VL \cite{bai2023qwen} + APS &  & 62.4 (\textbf{+0.4\%})  \\ \hline
 Monkey \cite{li2024monkey} & \multirow{2}{*}{896$\times$896} & 64.3  \\
Monkey \cite{li2024monkey} + APS & & 65.2 (\textbf{+1.4\%}) \\ \hline
\end{tabular}}
\end{table}

\subsection{Evaluation Metrics for Each Benchmark}
DocVQA \cite{mathew2021docvqa}, InfoVQA \cite{mathew2022infographicvqa}, ChartQA \cite{masry2022chartqa}, WTQ \cite{pasupat2015compositional}, and VisualMRC \cite{tanaka2021visualmrc} are all datasets designed for visual document image question answering. DocVQA encompasses a variety of document types, featuring a rich diversity of questions and data types. InfoVQA focuses on infographics that contain an abundance of visual elements. ChartQA and WTQ specifically target question answering for chart data and tabular data, respectively. VisualMRC, on the other hand, places a greater emphasis on developing natural language understanding and generation capabilities. DeepForm \cite{svetlichnaya2020deepform}, FUNSD \cite{jaume2019funsd}, SROIE \cite{huang2019icdar2019}, and POIE \cite{kuang2023visual} are key information extraction datasets where the model is required to identify the value corresponding to a given key.

Widely used text-centric benchmarks employ diverse metrics to better capture the distinct characteristics among various datasets. As depicted in Table 2 within the main text, both DocVQA \cite{mathew2021docvqa} and InfoVQA \cite{mathew2022infographicvqa} utilize ANLS \cite{biten2019icdar} as their metric. The results reported are evaluated on their respective official websites. ChartQA \cite{masry2022chartqa} is evaluated by accuracy relaxed accuracy \cite{methani2020plotqa}, while VisualMRC \cite{tanaka2021visualmrc} relies on CIDEr \cite{vedantam2015cider}. DeepForm \cite{svetlichnaya2020deepform} is measured by F1 score, and WTQ \cite{wtq} is evaluated by accuracy.

\subsection{APS's Performance for Scene Text Images}

APS is primarily designed for document images, where redundant regions are commonly distinctly defined. In contrast, elements in natural scene images tend to be more complex, making it challenging to identify extensive redundant regions. Consequently, APS generally has a limited impact on natural scene tasks. To validate this, we applied APS to the TextVQA dataset. As shown in Table \ref{tab:Adaptive Pixel Slimming supp}, APS does not introduce any negative impact. Some visualization results of APS in this scenario can be found in the last row of Fig. \ref{fig:Adaptive Pixel Slimming additional}.

\subsection{More Details and Results of Adaptive Pixel Slimming}
\begin{algorithm}[t]\footnotesize
\centering
\caption{Adaptive Pixel Slimming}
\label{al:Adaptive Pixel Slimming}
\begin{algorithmic}[1]
\Require
\Statex Source document image $I_S \in \mathbb{R}^{H\times W \times 3}$
\Statex SIZE=2048 
\Statex $t_n=50$
\Statex $t_c=10$
\Statex ICR(Identify continuous regions where the values exceed $t_v$ and the count of consecutive occurrences is greater than $t_c$.)
\State Sobel$_x$ = abs(Sobel($I_S$, direction=$x$))
\State Sobel$_y$ = abs(Sobel($I_S$, direction=$y$))
\State Initialize $I_G \in \mathbb{R}^{H\times W}$ 
\State $I_G(i,j)$ = max(Sobel$_x(i,j)$, Sobel$_y(i,j)$)
\State $I_G[I_G<t_n]=0$
\State $I_G$ = resize($I_G$, (SIZE, SIZE))
\State sum$_y$ = sum($I_G(i,j)$, along $y$ direction) $\in \mathbb{R}^{SIZE}$
\State blocks$_y$ = ICR(sum$_y$,$t_v$,$t_c$)
\State blocks$_y$ = normalize\_to\_original\_size(blocks$_y$, (H, W), SIZE)
\State sum$_x$ = sum($I_G(i,j)$, along $x$ direction) $\in \mathbb{R}^{SIZE}$
\State blocks$_x$ = ICR(sum$_x$,$t_v$,$t_c$)
\State blocks$_x$ = normalize\_to\_original\_size(blocks$_x$, (H, W), SIZE)
\State $I_t$ = remove\_blocks(blocks$_y$, blocks$_x$, $I_S$)
\State \Return $I_t$
\end{algorithmic}
\end{algorithm}

\begin{figure*}[]
  \centering
  \includegraphics[width=5in]{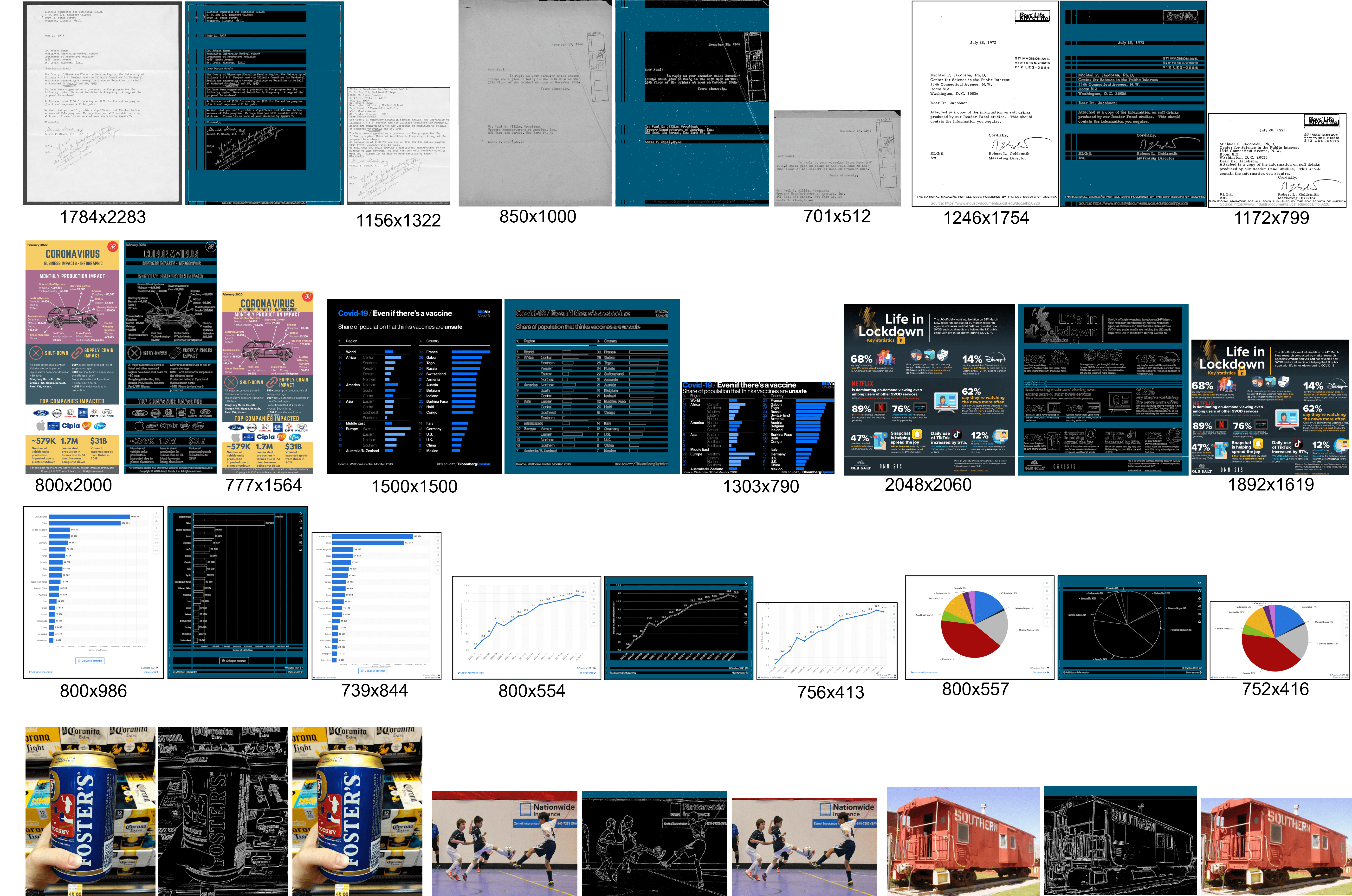}
  \caption{Additional visual results of Adaptive Pixel Slimming on various types of data. The redundant regions identified by Adaptive Pixel Slimming are highlighted on the corresponding gradient maps. The resolution for some images is indicated below the respective images.}
  \label{fig:Adaptive Pixel Slimming additional}
\end{figure*}

The specific computation method and parameter settings for Adaptive Pixel Slimming are detailed in Algorithm \ref{al:Adaptive Pixel Slimming}, which provides additional details beyond the main text: 1. The 2D gradient map is obtained by taking the maximum value from the gradient maps in the x and y directions. This approach aims to ensure the accuracy of redundancy detection while minimizing false positives. Using the maximum value rather than the average prevents the removal of regions with features in a single direction, such as table lines. 2. Noise removal is performed by setting gradient values below a threshold $t_n$ to zero, thereby reducing interference from background texture noise. 3. Before identifying redundant regions based on the value threshold $t_v$, we normalize the gradient map to a fixed resolution. This normalization ensures that the threshold is robust to varying resolutions. 

In Fig. \ref{fig:Adaptive Pixel Slimming additional}, we present some results of Adaptive Pixel Slimming, demonstrating its effectiveness in removing redundant regions from document images and reducing image resolution. The impact on natural scene images is minimal.

\subsection{More Details and Results of Dynamic Token Slimming}
The pseudocode for the Similarity Weighted Aggregation method in DTS is shown in Algorithm \ref{al:merging}.

We present the visualization results of APS and DTS on different data types in Figure 1. From these visualizations, we can draw the following conclusions: 1) APS effectively removes redundant regions when document backgrounds are simple, but struggles in scenarios involving graphics, document borders, table lines, or photographs. 2) In contrast, DTS is more robust and can handle these challenging cases that APS cannot. 3) While DTS is not perfect and may mistakenly classify some text areas as redundant, the impact of this issue in practical applications is minimal because: As detailed in the methodology section, DTS does not completely discard redundant tokens but retains as much information as possible through similarity-weighted aggregation; and although DTS might mask useful areas at the image level, during the image encoding process, information from these regions is integrated into neighboring tokens.

\begin{algorithm}[t]\footnotesize
\caption{Similarity Weighted Aggregation}
\label{al:merging}
\begin{algorithmic}[1]
\Require
\Statex essential\_tokens 
\Statex nonessential\_tokens 
\Statex FMS (find the most similar token)
\State mappings = \{\}
\State \textbf{for} token in nonessential\_tokens:
\State \hspace{0.3cm} most\_similar\_essential\_token = FMS(token, tokens in essential\_tokens )
\State \hspace{0.3cm} mappings[token] = most\_similar\_essential\_token 
\State updated\_essential\_tokens = []
\State \textbf{for} token in essential\_tokens:
\State \hspace{0.3cm} similar\_nonessential\_tokens = []
\State \hspace{0.3cm} \textbf{for} key,value in mappings:
\State \hspace{0.6cm} \textbf{if} token == value:
\State \hspace{0.9cm} similar\_nonessential\_tokens.append(key)
\State \hspace{0.2cm} \textbf{if} len(similar\_nonessential\_tokens)==0:
\State \hspace{0.6cm} updated\_essential\_tokens.append(token)
\State \hspace{0.2cm} \textbf{else}:
\State \hspace{0.6cm}  updated\_essential\_tokens.append(weighted\_sum(token, similar\_nonessential\_tokens))
\State \Return updated\_essential\_tokens
\end{algorithmic}
\end{algorithm}

\begin{figure*}[t]
  \centering
  \includegraphics[width=5.5in]{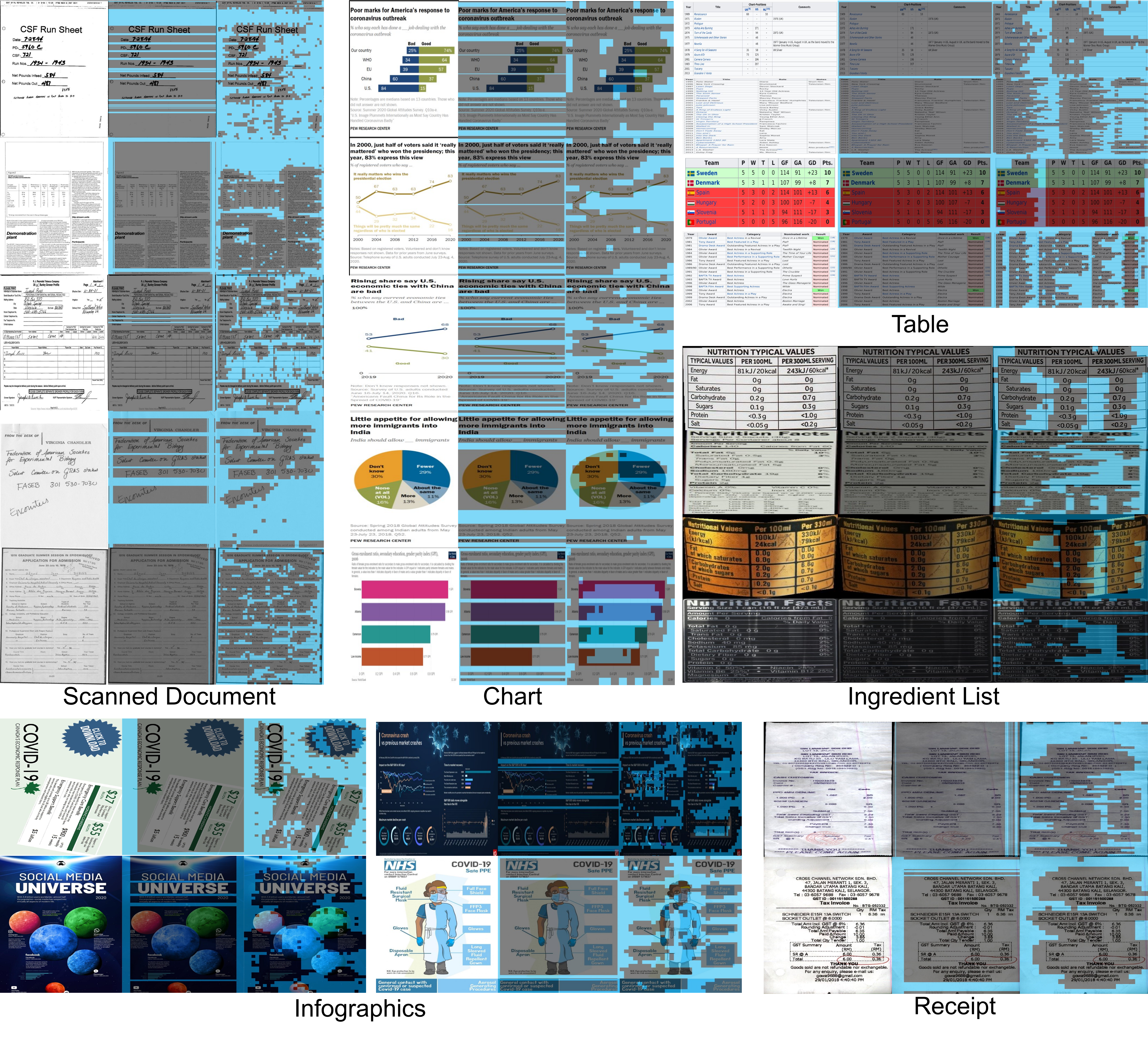}
  \caption{More visualized results of APS and DTS. The identified redundant regions are highlighted in light blue.}
  \label{fig:more_results}
\end{figure*}

\subsection{Edge Cases for DTS}
As shown in Algorithm \ref{al:clustering}, it is possible for DTS to encounter a situation where num1 equals num2. To investigate this issue, we re-evaluated the DocVQA dataset, recording values for num1 and num2 across 4335 samples. We found that the maximum, minimum, mean, and variance of num1 - num2 are 50, 0, 47, and 9, respectively (with num1 + num2 = 50). Only one sample had num1 - num2 = 0, and the proportion of samples with a difference of less than 10 was only 0.2\%. These indicate that num1 - num2 = 0 is a very rare case, with high confidence in most instances (large differences). Further analysis of this rare sample revealed a low input resolution with minimal redundant areas, leading some non-redundant tokens to be mistakenly included in the top 50 tokens by similarity. This issue can be resolved by adding a check to bypass DTS when the confidence is low (small difference). 

\end{document}